\pdfoutput=1
\documentclass[11pt]{article}

\usepackage[final]{acl}
\usepackage{booktabs}
\usepackage{placeins}
\usepackage{times}
\usepackage{latexsym}
\usepackage{float}
\usepackage[T1]{fontenc}
\usepackage{caption}
\usepackage{graphicx}
\usepackage{array} % 加载 array 包以使用 >{\bfseries} 等高级功能
\usepackage{tabularx}
\usepackage{fontawesome}
\usepackage{subfig}
\usepackage{float}
\usepackage{multirow}
\usepackage{arydshln} % For dashed lines
\usepackage{tcolorbox}
\definecolor{lightred}{RGB}{230, 145, 145}
\newtcolorbox{promptbox}[2][Prompt]{
colback=black!5!white,
arc=5pt, 
boxrule=0.5pt,
fonttitle=\bfseries,
title=#1, 
before upper={\small}, fontupper=\fontfamily{ptm}\selectfont,
colframe=#2,
}
\usepackage{xspace}
\newcommand{\bench}{\textit{LongIns}\xspace}
\usepackage[utf8]{inputenc}
\usepackage{natbib}
\usepackage{microtype}

\usepackage{inconsolata}

\usepackage{graphicx}

\title{\bench: A Challenging Long-context Instruction-based Exam \\for Large Language Models} %

\author{
    $^1$\textbf{Shawn Gavin}\footnotemark[1], 
    $^1$\,$^2$\,$^3$\textbf{Tuney Zheng}\footnotemark[1],
    $^1$\textbf{Jiaheng Liu},
    $^1$\textbf{Quehry Que},\\
    $^1$\,$^3$\textbf{Noah Wang},
    $^1$\textbf{Jian Yang},
    $^1$\textbf{Chenchen Zhang},\\
    $^3$\textbf{Wenhao Huang},
    $^1$\,$^2$\,$^3$\textbf{Ge Zhang}\footnotemark[2]\\
    $^1$M-A-P, $^2$University of Waterloo,
    $^3$01.ai
}

\begin{document}

\maketitle

\renewcommand{\thefootnote}{\fnsymbol{footnote}}
\footnotetext[1]{These authors contributed equally.}
\footnotetext[2]{Corresponding Authors.}
\renewcommand{\thefootnote}

\begin{abstract}
The long-context capabilities of large language models (LLMs) have been a hot topic in recent years. 
To evaluate LLMs' performance in different scenarios, various assessment benchmarks have emerged.
However, as most of these benchmarks focus on identifying key information to answer questions,
which mainly requires the retrieval ability of LLMs,
these benchmarks can partially represent the reasoning performance of LLMs from large amounts of information.
% it is limited by the typically short length of key information, allowing models to achieve high evaluations by focusing on only a small amount of information.
Meanwhile,
although LLMs often claim to have context windows of 32k, 128k, 200k, or even longer, these benchmarks fail to reveal the actual supported length of these LLMs. To address these issues, we propose the \textbf{\bench} benchmark dataset, a challenging long-context instruction-based exam for LLMs,
which is built based on the existing instruction datasets.
Specifically,
in our \bench,
we introduce three evaluation settings: \textbf{Global Instruction \& Single Task (GIST)}, \textbf{Local Instruction \& Single Task (LIST)}, and \textbf{Local Instruction \& Multiple Tasks (LIMT)}.
Based on \bench,
we perform comprehensive evaluations on existing LLMs and have the following important findings:
(1). The top-performing GPT-4 with 128k context length performs poorly on the evaluation context window of 16k in our \bench.
(2). For the multi-hop reasoning ability of many existing LLMs,
significant efforts are still needed under short context windows (<4k).

% a method for evaluating the true attentive length of LLMs through the in-context learning paradigm. This method allows us to determine the ``true comprehensible continuous maximum length'' under the ``claimed maximum reception length'' of different models.
\end{abstract}
\begin{figure*}[t]
  \centering
    \centering
    \includegraphics[width=\linewidth]{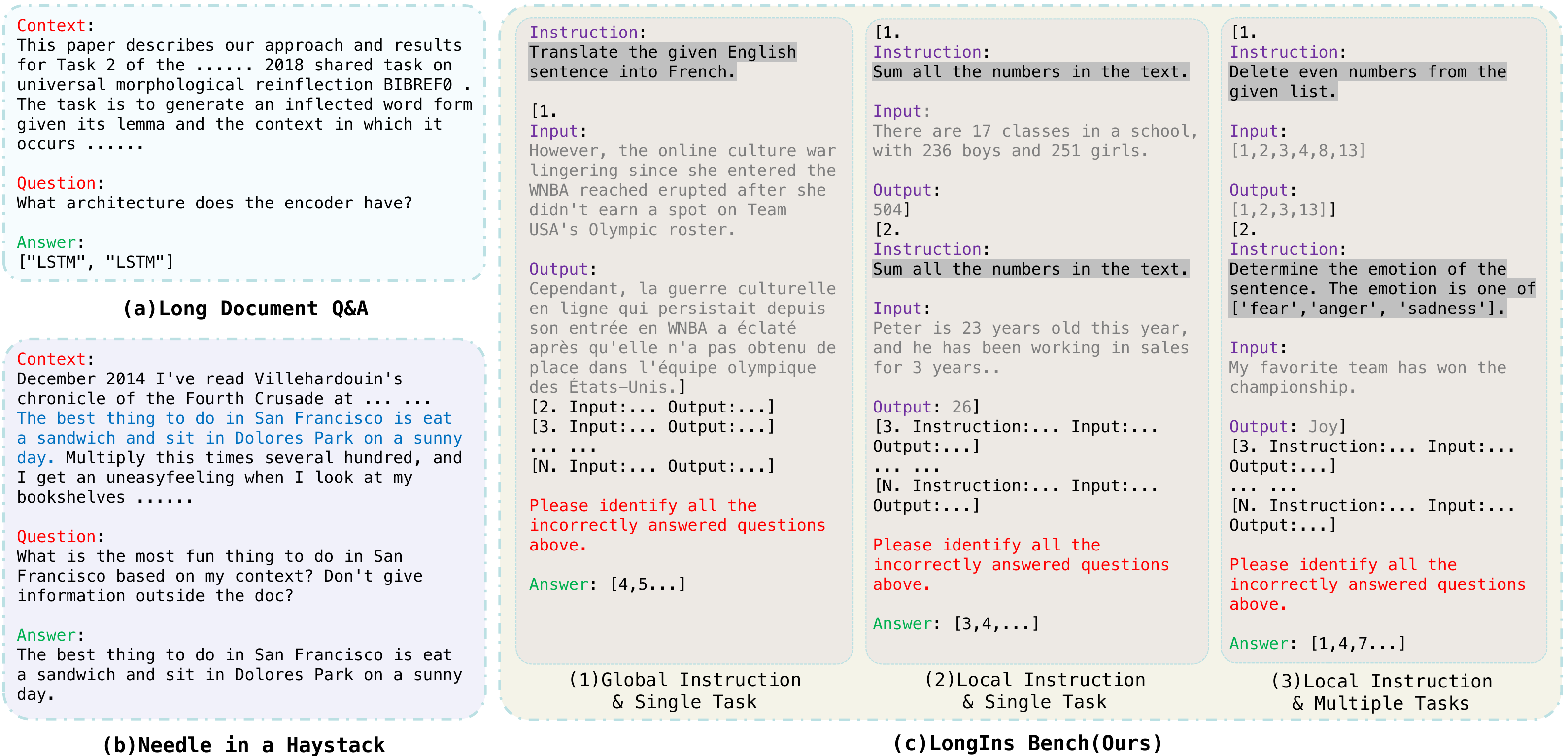}
  \hfill
    \centering
  \caption{(a) Long Document Q\&A is one of the common types of questions in mainstream long-context benchmarks. (b) Needle in a Haystack is an evaluation paradigm that tests the retrieval capabilities of LLMs in long contexts by inserting key information into long texts and asking questions based on that. (c) Our \bench consists of (1) GIST mode, Each sample is composed of the same task type concatenated together, and the task instruction is given only once at the beginning globally. (2) LIST mode, which is similar to GIST in the composition of samples, but instructions are provided for each question within the sample; (3) consists of questions from different task types concatenated together.}
  \label{fig:compare}
\end{figure*}

\section{Introduction}

The topic of extending the context window length in large language models (LLMs) remains a focal point in current research. Existing LLMs can handle context lengths ranging from 32k to 200k tokens \citep{wang2024beyond}, while some LLMs achieving capacities up to 10M tokens~\citep{liu2024world, bai2023qwen}. The capacity of length expansions is crucial for enhancing long document comprehension and forms the basis for a series of advanced applications, including repository-level code. Comprehension/Editing and the dependence of agent-based tasks on long-term memory. Despite these advancements, the predominant evaluation metrics continue to prioritize the retrieval performance of LLMs, overlooking the actual window length that the model can understand when faced with long texts.
%This oversight underscores the need for a new benchmark specifically designed to evaluate the long text processing abilities of LLMs comprehensively.

% The extension of context window length is consistently a hot topic in the field of large language models (LLMs). Existing LLMs support context window lengths of up to 32k-200k tokens ~\citep{liu2024world}, with some even reaching a maximum of 10M tokens~\citep{bai2023qwen}. Longer context windows not only enable models to perform long document reading comprehension but also serve as the foundation for various extended technical applications, such as code understanding and editing in repositories, or providing long historical memory in agent tasks. Despite the advancements in extending the context window length of LLMs, existing evaluation methods still focus on assessing the model's retrieval capabilities, neglecting the actual window length that the model can understand when faced with long texts. To address this significant gap, we advocate for developing a new benchmark to comprehensively evaluate the long text processing capabilities of these LLMs.

The capabilities of LLMs in scenarios involving long-context , such as cross-document aggregation, localization, and context tracking, is paramount for ensuring a seamless interaction between users and LLMs. Users often expect LLMs to read and understand the entire documents. However, existing benchmarks measure the proficiency of LLMs based on their capacity to retrieve and understand only key information from the inputs, thereby bypassing the necessity for an exhaustive understanding of the entire text. This does not align well with the real-world expectations from users who seek in-depth processing of long texts. Datasets commonly used to evaluate the long-text capabilities of LLMs, such as L-eval ~\citep{an2023leval} or longBench ~\citep{bai2023longbench}, are insufficient for assessing the understanding of long sequences. \textit{Therefore, how to truly evaluate the capabilities of LLMs in handling long-context tasks of the realistic scenarios still requires further exploration.}

To bridge this gap, we introduce \bench, a benchmark tailored to critically assess the proficiency of LLMs in understanding extensive sequences. \bench incorporates a contextual learning approach where the input involves more substantial key information segments, and is poised to ensure that correctly answering the culminating question necessitates a deep comprehension of the entire lengthy input sequence. This approach mandates that LLMs must truly excel in long-sequence understanding capabilities, beyond just the retrieval of key information from extended texts. We evaluate the performance of 20 different LLMs under \bench, including GPT-4o. We observe that LLMs generally perform worse on tasks requiring understanding of complete long sequences compared to retrieval tasks of the same length. As the total length increases, the performance gap becomes more pronounced. Additionally, the models' performance is largely independent of their advertised context window length, leading us to believe that the advertised context window length for most models should be understood as the "maximum acceptable sequence length" rather than the "maximum comprehensible sequence length."

Moreover, by controlling the distribution of incorrect answers and response situations, we analyze the distribution of attention changes at different text positions. Further analysis on the density of key information within the same total length shows that, except for GPT-4 and GPT-4o, the accuracy of most models rapidly declines as the density of key information increases. In summary, the primary contributions can be summarized as:
\begin{itemize}
\item We provide the \bench benchmark test, specifically designed to evaluate the actual understanding and processing capabilities of LLMs on long sequences. Unlike other benchmarks based on retrieval tasks, we focus more on assessing the actual comprehensible window length of the models.
\item In our \bench,
we introduce three evaluation settings: {Global Instruction \& Single Task (GIST)}, {Local Instruction \& Single Task (LIST)}, and {Local Instruction \& Multiple Tasks (LIMT)} to comprehensively evaluate the long-context abilities of existing LLMs.
\item We evaluate a series of long-context LLMs using this benchmark and find that most models fail to achieve high scores when the critical information length is only 8k. Even GPT-4 and GPT-4o score poorly at 16k length. This result is significantly different from the commonly recognized long context lengths (128k or longer), indicating that current LLMs still have considerable shortcomings in performing such tasks. We hope these results can provide a reference for research in the field of long-context LLMs.
\end{itemize}

\section{Related Work}
\label{gen_inst}

\paragraph{Long-context LLM}

The computational cost of processing sequences in Transformer-based models increases quadratically with sequence length, resulting in higher resource consumption and performance issues when handling long context inputs. Many studies explore various strategies to address this challenge, including the use of new positional encodings to achieve position interpolation or extrapolation~\cite{yarn,position-interpolation,liu20242}. For example, 
% FlashAttention enhances GPU utilization and achieves more efficient long-text training by leveraging the capability of seq\_len.
Rope~\citep{su2021roformer} extends the positional knowledge learned during the pre-training phase to longer sequence lengths through extrapolation, including various variants such as NTK-RoPE~\citep{ntk_rope_2021}. Alibi~\citep{press2021train}, on the other hand, maps long sequences to within recognizable lengths through interpolation. Some studies attempt to fine-tune LLMs to give the model a longer context window. LongLoRA ~\citep{chen2024longlora}is an efficient fine-tuning approach that significantly extends the context sizes of pre-trained LLMs with limited computational cost by using sparse local attention and improved parameter-efficient fine-tuning techniques. RingAttention~\citep{liu2023ring} can reduce the memory requirements of the Transformer model, allowing it to train sequences over 500 times longer than previous memory-efficient methods, without needing to approximate the attention mechanism. Additionally, traditional engineering techniques such as sliding windows or RAG~\citep{lewis2020retrieval} are also solutions for addressing long context scenarios in LLMs. These methods improve the performance of LLMs in handling long contexts in certain aspects.

\paragraph{Long-context Benchmark}
 To evaluate the performance of different LLMs on long texts, various benchmarks are often used to test different aspects of LLMs' capabilities. For instance, Zeroscrolls~\citep{shaham2023zeroscrolls} is a zero-shot benchmark for natural language understanding over long texts, which contains only test and small validation sets, without training data. LongBench~\citep{bai2023longbench} is a bilingual, multi-task benchmark for long context understanding, which also provides a subset with uniformly distributed data called LongBench-E.  L-Eval~\citep{an2023leval} constructs a long context evaluation benchmark containing multiple tasks and domains, including multiple-choice questions, with data that has undergone strict manual screening to ensure high quality. LooGLE~\citep{li2023loogle}, as a benchmark with longer samples, effective for evaluating the ability of LLMs to understand short-term and long-term dependencies in the content. M4LE~\citep{kwan_m4le_2023} provides additional tasks and datasets in both Chinese and English, covering a wider range of domains. $\infty$BENCH is the first benchmark with more than 100K tokens, with the longest test items reaching up to 2000K tokens. Needle in a Haystack~\citep{doe2022needle} employs Paul Graham's 218 essays as the ``haystack'' inserting a single sentence at various depths within the documents to statistically analyze the model's accuracy in identifying the locations of various ``needles'' and thereby evaluate its performance.

\begin{table}[htbp]
\centering
\small
\resizebox{\linewidth}{!}{%
\begin{tabular}{@{}lccc@{}}
\toprule
\textbf{Model} & \textbf{Size} & \textbf{Strategy} & \textbf{Support} \\ 
\midrule
GPT-4o & \faLock{} & \faLock{} & 128k  \\ 
GPT-4-Turbo & \faLock{} & \faLock{} & 128k \\
GPT-3.5-Turbo & \faLock{} & \faLock{} & 16K \\ 
ERNIE-Speed & \faLock{} & \faLock{} & 8k  \\ 
Qwen-Long & \faLock{} & NTK+LNS+WAttn & 10M \\ 
Deepseek-Chat & \faLock{} & YaRN & 32k  \\ 
Moonshot-v1 & \faLock{} & \faLock{} & 128k \\ 
Yi-Large-Turbo & \faLock{} & SeqPara+DistrAttn & 200k \\ 
Yi-Spark & \faLock{} & SeqPara+DistrAttn & 200k \\ 
GLM-4 & \faLock{} & \faLock{} & 128k \\ [2pt]\hdashline\\[-8pt]
ChatGLM2-6B & 6B & PI & 32k \\
Deepseek-LLM-Chat & 7B & - & 4k\\
Lwm-Text-Chat & 7B & RingAttn & 128k \\
LongChat-7B & 7B & Rope & 16k \\
MAP-Neo-Ins-v0.1 & 7B & - & 8k \\
Mistral-7B-Ins-v0.2 & 7B & Rope & 32k\\
Longpalca-7B & 7B & LongLoRA & 32k \\
Qwen1.5-7B-Chat & 7B & Rope+Sliding Window & 32k \\
Internlm2-Chat-7B & 7B & Dynamic NTK & 200k \\
Llama3-8B-Ins & 8B & RoPE & 8K\\ [2pt]\hdashline\\[-8pt]
LongChat-13B & 13B & Rope & 16k\\
Baichuan2-13B-Chat & 13B & - & 4k \\ 
\bottomrule
\end{tabular}
 }
\caption{Details on the evaluated models. ``Ins'' indicates ``Instruct''. ``LNS'' indicates ``LogNScaling''. ``WAttn'' indicates ``WindowAttention''}
\label{tab:model_info}
\end{table}

% \begin{itemize}
% \item{}\textbf{SNI}~\citep{wang-etal-2022-supernaturalinstructions} (Super-NaturalInstructions) dataset aims to enhance the generalization ability of NLP models to unknown tasks through natural language instructions. SNI contains over 1,600 NLP tasks and instructions, covering various task types such as classification, extraction, filling, and sequence labeling. This dataset emphasizes the use of declarative instructions to improve the model's generalization ability and extends the previous Natural Instructions dataset. It allows for the evaluation and comparison of different NLP models in understanding and executing these task instructions.
% \item{}\textbf{BIG-bench}~\citep{bigbench} (Broad Interactive General Benchmark) is a multi-task dataset designed to evaluate and compare the performance of large language models. It consists of multiple independent sub-tasks covering a wide range of topics and question types, including comprehension, reasoning, translation, arithmetic, and more. The goal of BIG-bench is to test the capabilities of language models on various complex and challenging tasks, particularly those that go beyond simple language understanding. A key feature of BIG-bench is its inclusion of tasks that are uncommon in traditional NLP benchmarks, such as humor detection, poetry composition, and logic puzzles. This approach aims to explore and challenge the performance limits of current language models in a broader and more diverse set of environments.
% \end{itemize}
\begin{table*}[htbp]
\centering
\resizebox{\linewidth}{!}{%
\begin{tabular}{@{}lcccccc@{}}
\toprule
 & \multicolumn{2}{c}{\textbf{GIST}} & \multicolumn{2}{c}{\textbf{LIST}} & \multicolumn{2}{c}{\textbf{LIMT}} \\
\cmidrule(lr){2-3} \cmidrule(lr){4-5} \cmidrule(lr){6-7}
 & \textbf{Number of Questions} & \textbf{Q-Density} & \textbf{Number of Questions} & \textbf{Q-Density} & \textbf{Number of Questions} & \textbf{Q-Density}\\
\midrule
QA  & 265 $\times$ 7 & 1.40  & 265 $\times$ 7 & 1.39  & - & -\\
Classif  & 229 $\times$ 7 & 1.67   & 229 $\times$ 7 & 1.65  & - & -  \\
RC & 142 $\times$ 7 & 1.73  & 142 $\times$ 7 & 1.71   & - & - \\
NLI  & 140 $\times$ 7 & 1.72 & 140 $\times$ 7 & 1.70   & - & - \\
MT & 432 $\times$ 7 & 1.10  & 432 $\times$ 7 & 1.09   & - & - \\
NER & 96 $\times$ 7 & 2.32   & 96 $\times$ 7 & 2.27   & - & - \\
CSR & 105 $\times$ 7 & 2.69  & 105 $\times$ 7 & 2.64   & - & - \\
\midrule
Total & 1409 $\times$ 7 & 1.41  & 1409 $\times$ 7 & 1.39 & 50 $\times$ 7 & 1.40 \\
\bottomrule
\end{tabular}
}
\caption{The distribution of task types in the LongBBH dataset. ``QA'', ``Classif'', ``RC'', ``NLI'', ``MT'', ``NER'', ``CSR'' represent Question Answering, Classification, Reading Comprehension, Natural Language Inference, Machine Translation, Named Entity Recognition, and Common Sense Reasoning, respectively. ``Q-Density'' represents the number of questions per 100 tokens, reflecting question density in each paper. Samples in LIMT consist of a mix of various task types, so the information is not presented by category.}
\label{tab:dataset_info}
\vspace{-0.2cm}
\end{table*}

\section{\bench Benchmark}

To support the evaluation of various types and lengths of tasks, we collect a diverse range of questions from the Super-NaturalInstructions (SNI) and BIG-bench datasets, covering different areas of natural language processing. This ensures comprehensive coverage of task types and difficulty levels. To address the issue of insufficient samples in some categories, we generate additional questions. As detailed in \autoref{tab:dataset_info}, our dataset, named \bench, includes seven different context lengths: 256, 512, 1024, 2048, 4096, 8192, and 16384 tokens, with 1409 questions for each length. These questions divide into seven task types, including question answering, classification, reading comprehension, natural language inference, translation, named entity recognition, and commonsense reasoning.

Specifically, we construct our dataset by concatenating questions to the specified context lengths and modifying some of the questions' answers to incorrect ones (we control the overall error rate at around 10\% but ensure at least one incorrect answer for shorter lengths). The aim is to reflect the long-context performance of LLMs based on their efficiency in identifying incorrect answers after reading multiple questions simultaneously. Based on the above concept, we construct a dataset using the Global Instruction \& Single Task (GIST) mode. Additionally, to explore the impact of different factors on the evaluation results, we create two other datasets: Local Instruction \& Single Task (LIST) and Local Instruction \& Multiple Tasks (LIMT).

\paragraph{Global Instruction \& Single Task}
The GIST dataset serves as our core benchmark. GIST evaluates by concatenating the same type of questions to the specified lengths, constructing prompts for each sample according to the template shown in \autoref{fig:prompt}. To assess the model's attention capabilities across different lengths and positions, we ensure an even distribution of incorrect answers to maintain statistical significance. During testing, for open-source models, we use the official example system prompt; if no official example is provided, we default to standard configurations. For closed-source models, we utilize examples from the official documentation.

\paragraph{Local Instruction \& Multiple Tasks}
Each piece of sample in GIST is composed of questions of the same task type, sharing a common task description. Therefore, the task description (i.e., Instruction) is placed at the beginning globally. However, this positioning might result in the subsequent questions being far from the Instruction, potentially affecting accuracy. To explore this issue, we construct a dataset in LIST mode for ablation experiments, as illustrated in \autoref{fig:prompt_local}.

\paragraph{Local Instruction \& Multiple Tasks}
Both GIST and LIST benchmarks are composed of questions of the same task type. When the context content is similar, the model's in-context learning ability becomes more significant. To explore the role of the model's in-context learning ability in the evaluation results, we design a subset called LIMT. LIMT includes 7 lengths, with each length containing 200 test items formed by mixing various task types. The construction template prompt for LIMT is similar to \autoref{fig:prompt_local}, with the only difference being that each question within the sample corresponds to a different instruction.

\section{Experiment}
In this chapter, we present the experimental methods as well as the main findings and results obtained through these experiments.
\begin{table*}[htbp]
\centering
\small
\resizebox{\linewidth}{!}{%
\begin{tabularx}{\textwidth}{@{} l *{8}{X} c @{}} % l 表示左对齐的固定列，*{6}{X}表示六个自适应宽度的列，c表示一个中心对齐的固定列
% \begin{tabular}{@{}lccccccc@{}}
\toprule
\textbf{Model} & \textbf{Size} &\textbf{Support}  & \textbf{256} & \textbf{512} & \textbf{1024} & \textbf{2048} & \textbf{4096} & \textbf{8192} & \textbf{16384} \\ 
\midrule
GPT-4o & \faLock{} & 128k &70.94 & 59.14 & 60.58 & 55.43 & 52.92 & 43.81 & 31.53 \\ 
GPT-4-Turbo & \faLock{} & 128k & 69.59 & 63.13 & 64.21 & 59.08 & 57.52 & 50.73 & 40.91 \\
GPT-3.5-Turbo & \faLock{} & 16K & 54.61 & 45.38 & 41.68 & 34.81 & 26.27 & 18.81 & 12.23 \\ 
ERNIE-Speed & \faLock{} & 8k & 44.16 & 35.99 & 24.88 & 17.29 & 12.22 & 1.13 & - \\ 
Qwen-Long & \faLock{} & 10M & 61.58 & 54.60 & 42.07 & 34.24 & 25.99 & 18.71 & 10.33 \\ 
Deepseek-Chat & \faLock{} &  32k & 45.71 & 40.86 & 32.41 & 19.32 & 11.33 & 5.42 & 3.10\\ 
Moonshot-v1 & \faLock{} &  128k & 69.45 & 61.22 & 52.35 & 52.94 & 34.89 & 24.29 & 15.10 \\ 
Yi-Large-Turbo & \faLock{} &  200k & 57.05 & 42.22 & 33.81 & 28.60 & 19.82 & 11.69 & 6.01 \\ 
Yi-Spark & \faLock{} &  200k & 51.61 & 39.04 & 35.53 & 24.44 & 15.97 & 5.08 & 0.88 \\ 
GLM-4 & \faLock{} &  128k & 69.20 & 64.66 & 56.62 & 56.11 & 33.50 & 23.72 & 12.01 \\  [2pt]\hdashline\\[-8pt]
ChatGLM2-6B & 6B & 32k & 20.75 & 17.00 & 14.08 & 10.81 & 6.38 & 2.93 & 0.87 \\
Deepseek-LLM-Chat & 7B & 4k & 29.08 & 24.01 & 16.75 & 12.99 & 3.61 & - & - \\   
Lwm-Text-Chat  & 7B & 128k & 22.19 & 15.20 & 10.77 & 8.80 & 3.18 & 0.73 & - \\
LongChat-7B  & 7B & 16k & 24.58 & 20.85 & 17.48 & 14.35 & 9.62 & 6.31 & 2.54 \\ 
MAP-Neo-Ins-v0.1 & 7B & 8k & 41.22 & 34.17 & 28.09 & 21.96 & 14.72 & 3.28 & -  \\ 
Mistral-7B-Ins-v0.2  & 7B & 32k & 40.53 & 37.00 & 29.15 & 20.37 & 0 & 0 & 0 \\
Longalpaca-7B  & 7B & 32k & 31.75 & 25.37 & 20.08 & 14.94 & 9.68 & 4.74 & 1.95 \\ 
Qwen1.5-7B-Chat  & 7B & 32k & 33.48 & 29.88 & 24.32 & 19.50 & 16.72 & 11.57 & 5.22 \\
Internlm2-Chat-7B  & 7B & 200k & 52.14 & 45.23 & 36.84 & 26.88 & 18.19 & 11.98 & 6.06 \\
Llama3-8B-Ins  & 8B & 8k & 52.93 & 46.35 & 40.63 & 32.06 & 22.48 & 10.09 & - \\  [2pt]\hdashline\\[-8pt]
LongChat-13B  & 13B & 16k & 27.96 & 25.52 & 22.19 & 18.06 & 14.30 & 8.21 & 3.15 \\ 
Baichuan2-13B-Chat  & 13B & 4k & 27.99 & 23.96 & 21.37 & 17.19 & 0 & - & - \\
Yi-34B  & 34B & 200k & 26.98 & 22.14 & 17.00 & 13.24 & 8.13 & 4.02 & 1.19 \\
\bottomrule
\end{tabularx}
}
\caption{The evaluation results on GIST Subset. We report scores of multiple open-source and closed-source models across different lengths. ``Ins'' indicates ``Instruct''. }
\label{tab:main-result}
\vspace{-0.2cm}
\end{table*}
\subsection{Experimental Setup}

In this paper, we conduct a comprehensive evaluation of recent open-source and closed-source LLMs~\citep{bai2023qwen,ai2024yi,zeng2022glm,deepseekai2024deepseekv2,du2022glm,liu2024world,longchat2023,zhang2024mapneo,2023internlm,llama3modelcard,baichuan2023baichuan2,achiam2023gpt,Moonshot,ERNIE}. Since the core idea of this benchmark is ``long key information'' rather than the total length of the text, we include some models that only retain the original 4k length window. 
\autoref{tab:model_info} provides basic information about the models we include.

The goal of this work is to construct a benchmark where LLMs must read and understand the entire context word by word to answer correctly. To achieve this, we employ an examination method requiring LLMs to carefully read all questions in a long text to correctly identify the ``numbers of questions with incorrect answers.'' A specific example can be found in \autoref{tab:example_prompt} in the appendix.

We also conduct experiments on the accuracy distribution of LLMs when incorrect questions are located at different depths for different text lengths. This approach reflects focus ability of LLMs at various positions and lengths. Additionally, since the length of questions affects LLM performance for the same total length, we conduct further analysis on question density under the same total length.

% \begin{figure*}[t]
%   \centering
%     \centering
%     \includegraphics[width=0.8\linewidth]{image/prompt.pdf}
%   \hfill
%     \centering
%   \caption{The prompt template used for evaluation in LongBBH.}
%   \label{fig:prompt}
% \end{figure*}

% The overall evaluation results of this paper are presented in \autoref{tab:main-result}. Our input consists of exam papers formed by concatenating multiple questions of the same type, with some of the answers intentionally altered to be incorrect. We generated prompt words using appropriate templates to expect the model, acting as a grading model, to identify all incorrectly answered questions. On this basis, we used the F1 score between the list of actual incorrect question numbers and the list of predicted incorrect question numbers by LLMs as an evaluation metric. From the results in the table, it can be seen that closed-source models generally score higher than open-source models, especially in longer key information fields. Notably, GPT-4-turbo and GPT-4o maintain scores of 40.91 and 31.53, respectively, even with an exam length of 16k tokens, whereas other models are practically unusable at this length. A 256-token exam length is considered relatively short in the context of large language model interactions, yet many LLMs still perform poorly at this length. This indicates that LLM performance declines when faced with slightly longer texts requiring continuous attention to key information. This reveals certain shortcomings in current LLMs, particularly in multi-hop compositional reasoning abilities.

\subsection{Results on GIST Subset} 
The overall evaluation results of our study are presented in \autoref{tab:main-result}. 
Our inputs are sample created by concatenating multiple questions of the same type, with some answers intentionally altered to be incorrect. 
We generate prompt words using appropriate templates to guide the model, which serves as a grading tool, to identify all incorrectly answered questions. 
We use the F1 score between the list of actual incorrect question numbers and predicted incorrect question numbers predicted by  LLMs.
﻿
The results in the table indicate that closed-source models generally outperform open-source models, especially in fields requiring longer key information. 
Notably, GPT-4-turbo and GPT-4o maintain scores of 40.91 and 31.53, respectively, even with an exam length of 16,000 tokens, whereas other models are practically unusable at this length. 
Although a 256-token exam length is considered relatively short in the context of LLM interactions, certain LLMs still perform poorly at this length. 
This suggests that LLM performance declines when facing slightly longer texts that require sustained attention to key information. These findings highlight certain shortcomings in current LLMs, particularly in their multi-hop compositional reasoning abilities.

\begin{table*}[htbp]
\centering
\small
\resizebox{\linewidth}{!}{%
\begin{tabularx}{\textwidth}{@{} l *{8}{X} c @{}} % l 表示左对齐的固定列，*{6}{X}表示六个自适应宽度的列，c表示一个中心对齐的固定列
% \begin{tabular}{@{}lccccccc@{}}
\toprule
\textbf{Model} & \textbf{Size} &\textbf{Support}  & \textbf{256} & \textbf{512} & \textbf{1024} & \textbf{2048} & \textbf{4096} & \textbf{8192} & \textbf{16384} \\ 
\midrule
GPT-4o & \faLock{} & 128k & 76.31 & 74.65 & 71.16 & 66.55 & 58.83 & 56.23 & 51.15 \\ 
GPT-4-Turbo & \faLock{} & 128k & 73.89  & 69.01 & 65.16 & 60.22 & 59.63 & 51.49 & 44.22 \\
GPT-3.5-Turbo & \faLock{} & 16K & 51.60 & 50.12 & 42.04 & 32.07 & 18.58   & 19.67 & 7.64 \\ 
ERNIE-Speed & \faLock{} & 8k & 42.34  & 36.77  & 33.52 & 23.67 & 16.86 & 4.18 & - \\ 
Qwen-Long & \faLock{} & 10M &  59.00 & 54.45  & 53.87 & 47.34  & 38.29 & 35.22  & 23.97 \\ 
Deepseek-Chat & \faLock{} &  32k & 69.08  & 67.56 & 59.19 & 47.28  & 44.67 & 41.56  & 35.23\\ 
Yi-Large-Turbo & \faLock{} &  200k & 50.03  & 44.43 & 38.53& 33.89 & 27.43 & 26.19 & 17.38 \\ 
Yi-Spark & \faLock{} &  200k & 43.93 & 38.19 &34.39 &  29.86 &  26.66 & 21.26 & 15.92 \\ 
GLM-4 & \faLock{} &  128k &  58.52 & 53.00  &48.58   &  44.64 & 42.15 & 41.02 & 38.74 \\  [2pt]\hdashline\\[-8pt]
MAP-Neo-Ins-v0.1 & 7B & 8k & 41.96   & 36.95  &  27.87  &  23.02   & 16.04  &  6.37 & - \\
ChatGLM2-6B & 6B & 32k & 20.75 & 17.00 & 14.08 & 10.81 & 6.38 & 2.93 & 0.87 \\
Deepseek-LLM-Chat & 7B & 4k & 29.08 & 24.01 & 16.75 & 12.99 & 3.61 & - & - \\ 
Mistral-7B-Ins-v0.2  & 7B & 32k & 40.53 & 37.00 & 29.15 & 20.37 & 0 & 0 & 0 \\
Qwen1.5-7B-Chat  & 7B & 32k & 33.48 & 29.88 & 24.32 & 19.50 & 16.72 & 11.57 & 5.22 \\
Internlm2-Chat-7B  & 7B & 200k & 52.14 & 45.23 & 36.84 & 26.88 & 18.19 & 11.98 & 6.06 \\
Llama3-8B-Ins  & 8B & 8k & 52.93 & 46.35 & 40.63 & 32.06 & 22.48 & 10.09 & - \\  
[2pt]\hdashline\\[-8pt]
LongChat-13B  & 13B & 16k & 29.02 & 25.67 & 22.58 & 19.26 & 12.97 & 9.31 & 5.51 \\ 
Baichuan2-13B-Chat  & 13B & 4k & 30.77 & 25.10 & 23.95 & 20.52 & 0 & - & - \\
\bottomrule
\end{tabularx}
}
\caption{The result of Local Instruction \& Multiple Tasks. ``Ins'' indicates ``Instruct''}
\label{tab:local mixed instructions}
\end{table*}
\subsection{Results on LIST Subset} 
\label{sec:local-single}
To investigate the effect of LLMs' dependence on instruction positioning, we conduct an ablation experiment by changing the instructions from global instructions to local instructions. The specific prompts are shown in \autoref{fig:prompt_local}. 
For the GIST subset, the instructions are given only at the beginning, but for the LIST subset, instructions are provided before each question to explore the model's dependency performance at different distances between the instructions and inputs. 
﻿
The results, as shown in \autoref{tab:local single instructions}, indicate that compared to the evaluation results on the GIST subset, models generally achieve higher scores on the LIST subset. 
Changing the position of instructions leads to higher model scores, indicating that most models are sensitive to the distance of instruction dependency. 
When instructions and inputs are separated by a greater distance, the performance of most models rapidly degrades. 
This is common across various models such as GPT-4-Turbo, including those that perform well in other tasks. 
And this issue generally affects the model's reliance on the early memories in multi-turn conversations. 
But as a trivial solution, repeating the important instructions would significantly increase inference costs.

\subsection{Results on LIMT Subset} 
To explore the long-context comprehension ability of LLMs outside the in-context learning paradigm, we also concatenate the questions of different task types to the same length and evaluate on them. 
We extract a mini-mix dataset comprising 50 samples for each of the 7 task types, totaling 350 samples. 
The evaluation results are shown in \autoref{tab:local mixed instructions}.

We observe that models generally score higher under the LIMT setting compared to the GIST setting. 
This is because questions from different types of tasks have different instructions, and therefore each output needs to be given separately.
As a result, the distance between the question and the instructions is closer, and the dependence on the instructions becomes more apparent, leading to higher scores.
In contrast, when compared with the evaluation results in \autoref{tab:local single instructions}, it is evident that the LIMT setting are more challenging than the LIST setting under the same input format. 
This indicates that most models still fall short in tasks outside of the in-context learning paradigm.

\section{Discussion}

During the analysis, we discover several noteworthy phenomena:
\begin{itemize}
\item{}The Yi-spark model, small-parameter model, achieves results close to those of larger parameter models, suggesting that the ability to focus on long key information is not strictly dependent on the number of parameters.
\item{}The ERNIE-Speed model has a relatively low score because its strict review mechanism results in many refusals to answer.
\item{}Some open-source models, such as baichuan2-13B-Chat and mistral-7b-32k, exhibit a sharp degradation when the length of key information approaches but does not exceed their context window limit. For instance, when the test length jumps from 2k to 4k tokens, these models suddenly fail to follow instructions correctly and start producing nonsensical outputs. We believe this implies the true length that some models can handle effectively.
\end{itemize}

\subsection{Effect of the Position of Key Information}

``Needle in a Haystack'' ~~\citep{doe2022needle} and ``Lost In the Middle'' ~~\citep{Liu2023LostIT} indicate LLMs exhibit varying levels of attention when processing different positions within long texts. Therefore, we analyze the LLMs' ability to handle key information located at different positions by controlling the distribution of error question.

\begin{figure}[H]
    \centering
    \begin{minipage}{0.235\textwidth}
        \centering
        \includegraphics[width=\textwidth]{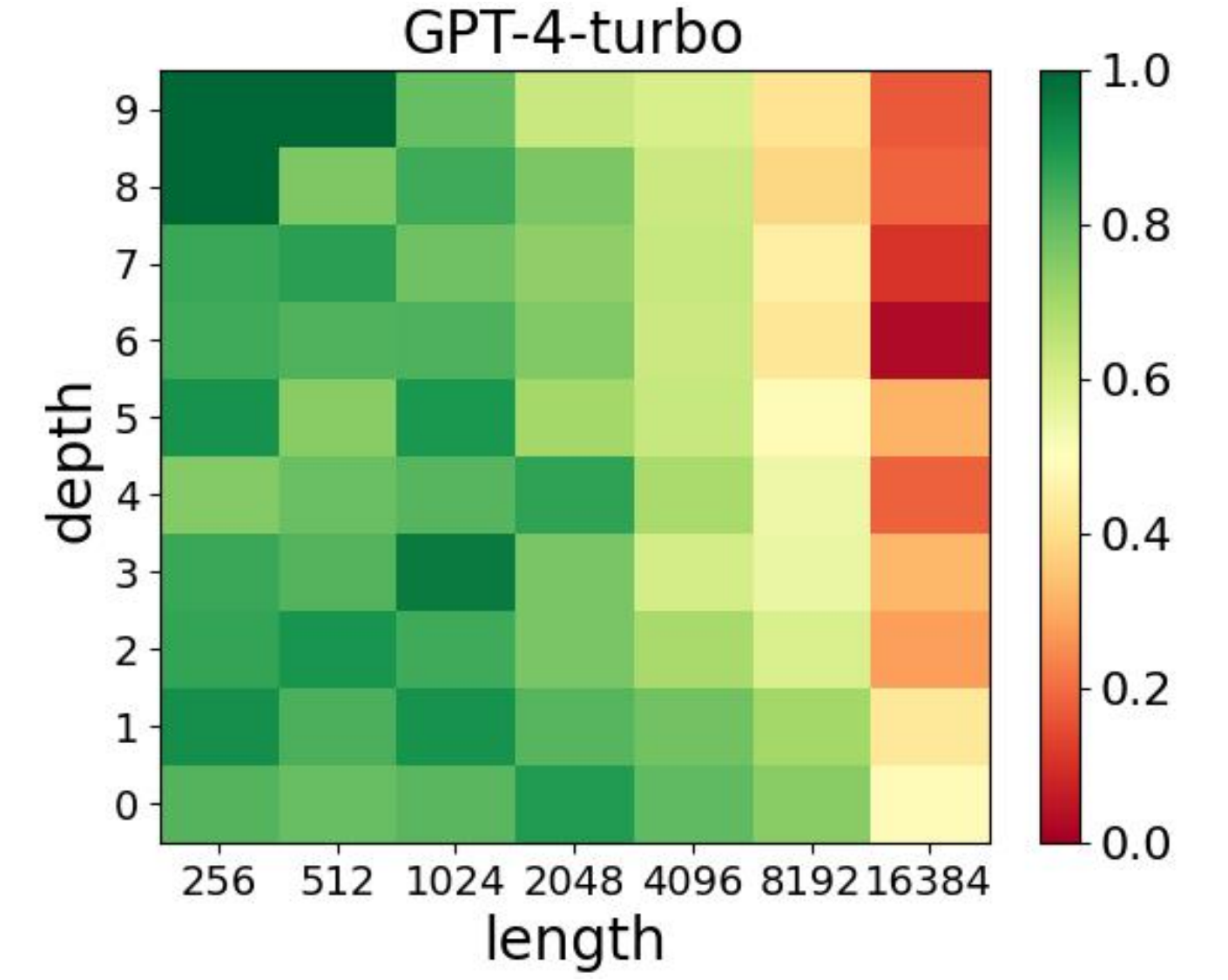}
        \label{fig:image1}
    \end{minipage}
    \hfill
    \begin{minipage}{0.235\textwidth}
        \centering
        \includegraphics[width=\textwidth]{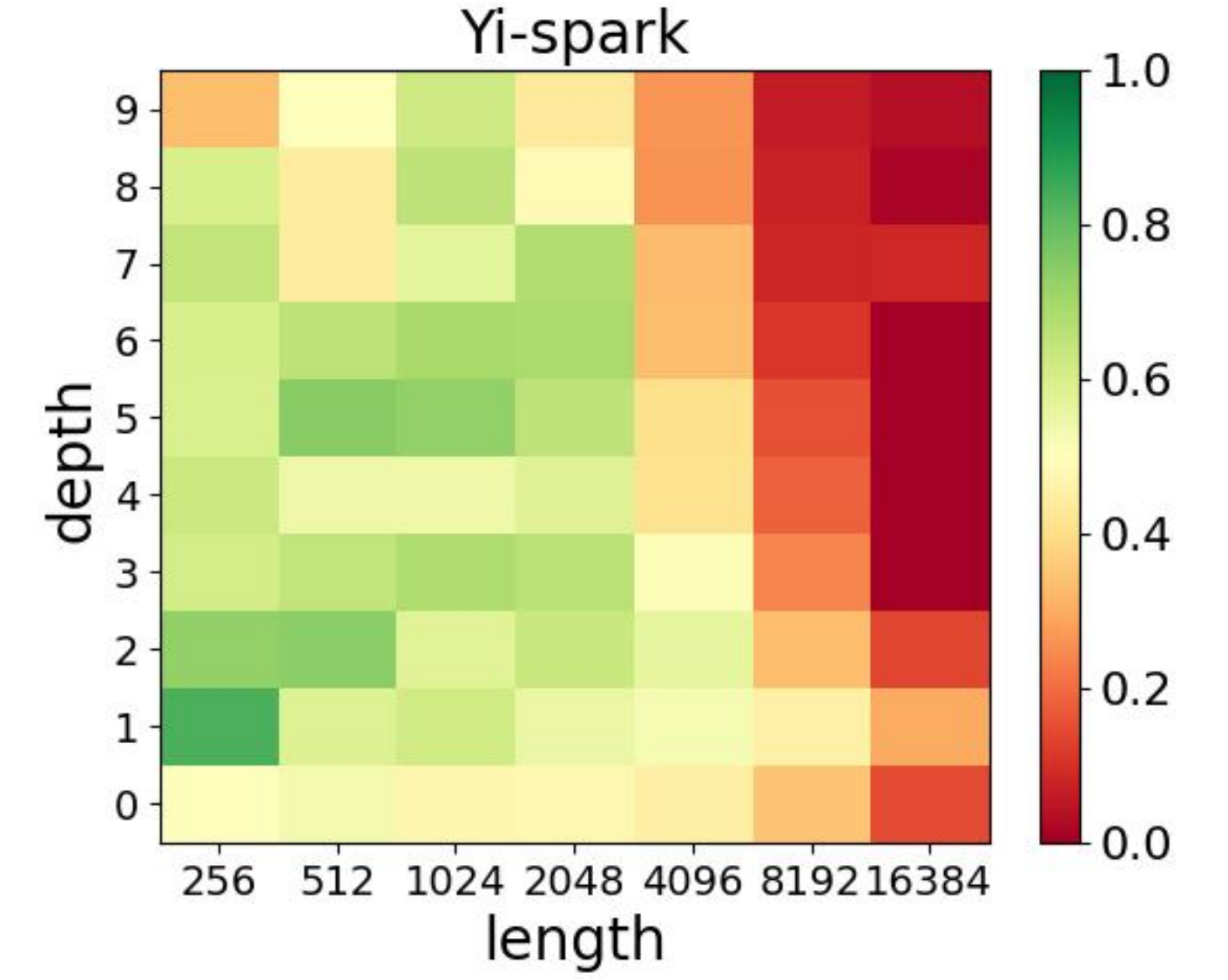}
        \label{fig:image2}
    \end{minipage}
    \vspace{0.1cm} % Adds vertical space between the rows
    \begin{minipage}{0.235\textwidth}
        \centering
        \includegraphics[width=\textwidth]{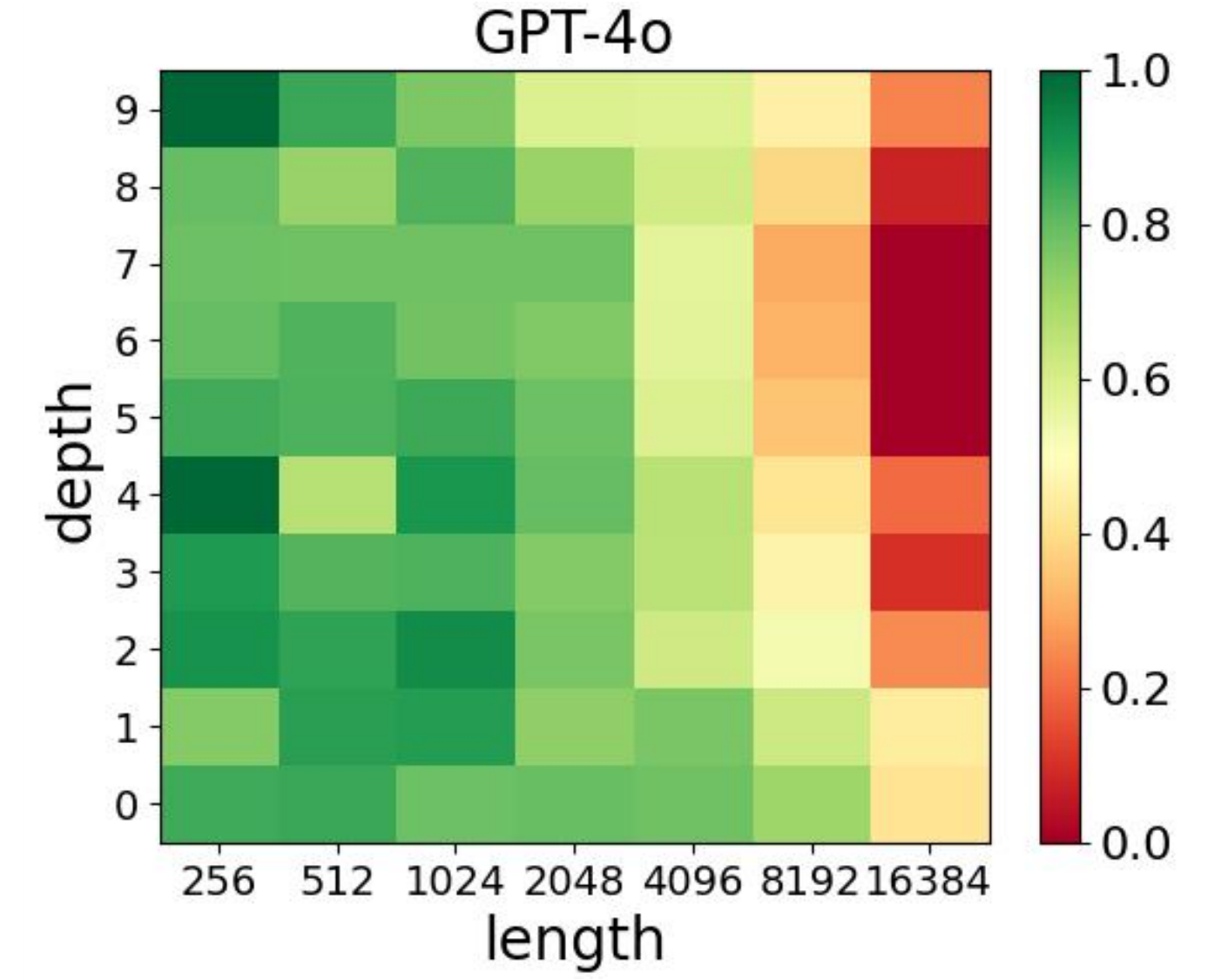}
        \label{fig:image3}
    \end{minipage}
    \hfill
    \begin{minipage}{0.235\textwidth}
        \centering
        \includegraphics[width=\textwidth]{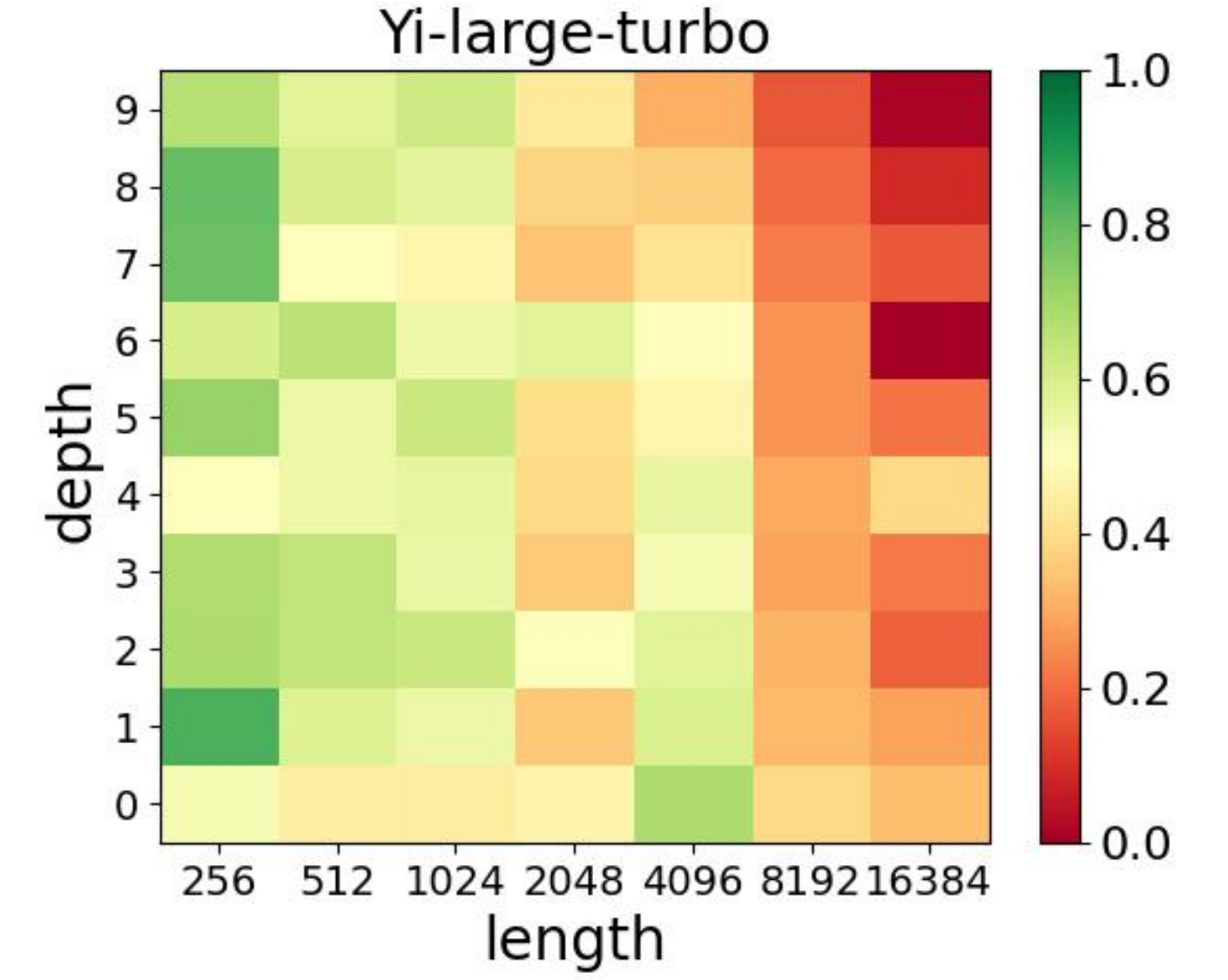}
        \label{fig:image4}
    \end{minipage}
    % \vspace{0.1cm} % Adds vertical space between the rows
    \begin{minipage}{0.235\textwidth}
        \centering
        \includegraphics[width=\textwidth]{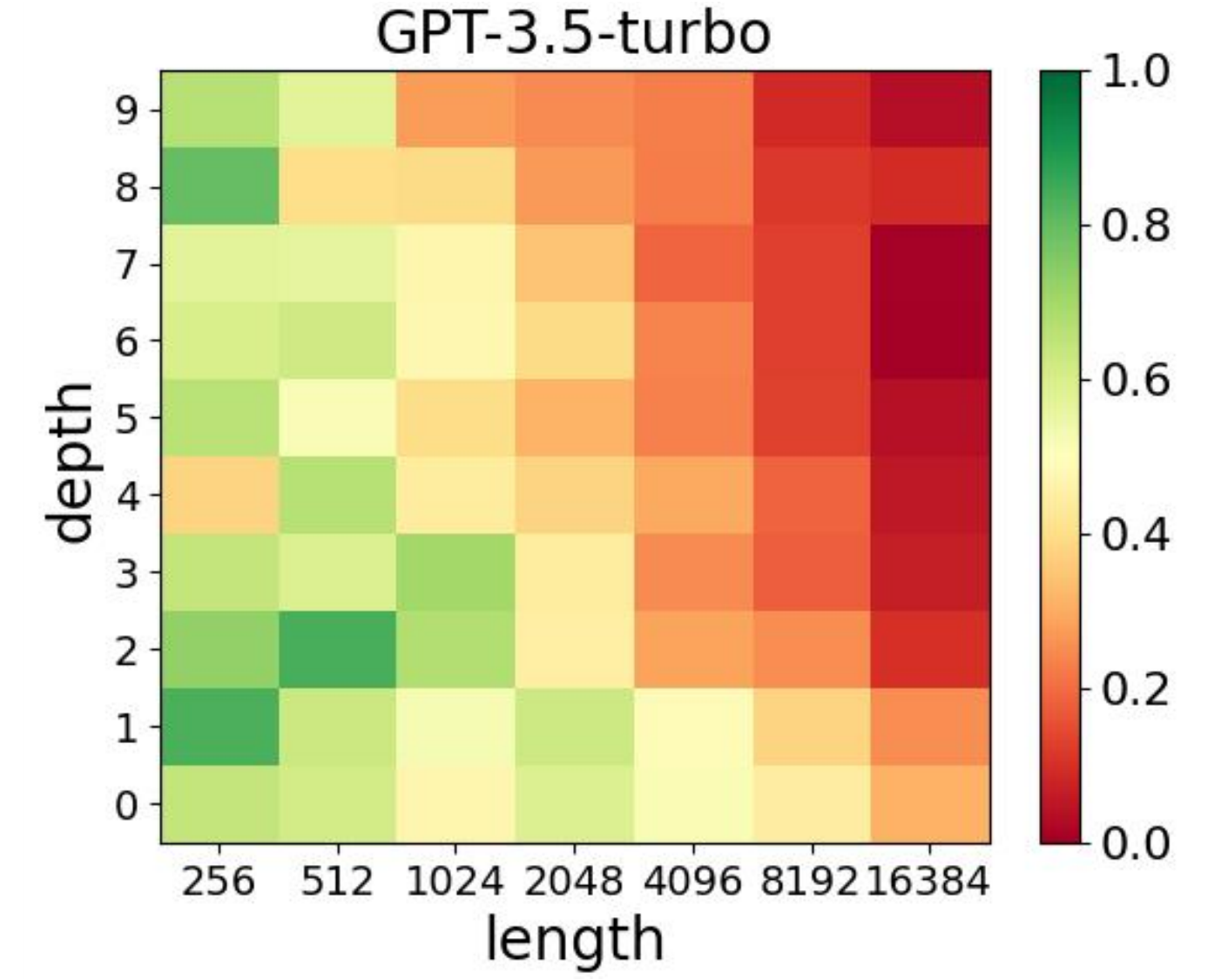}
        \label{fig:image5}
    \end{minipage}
    \hfill
    \begin{minipage}{0.235\textwidth}
        \centering
        \includegraphics[width=\textwidth]{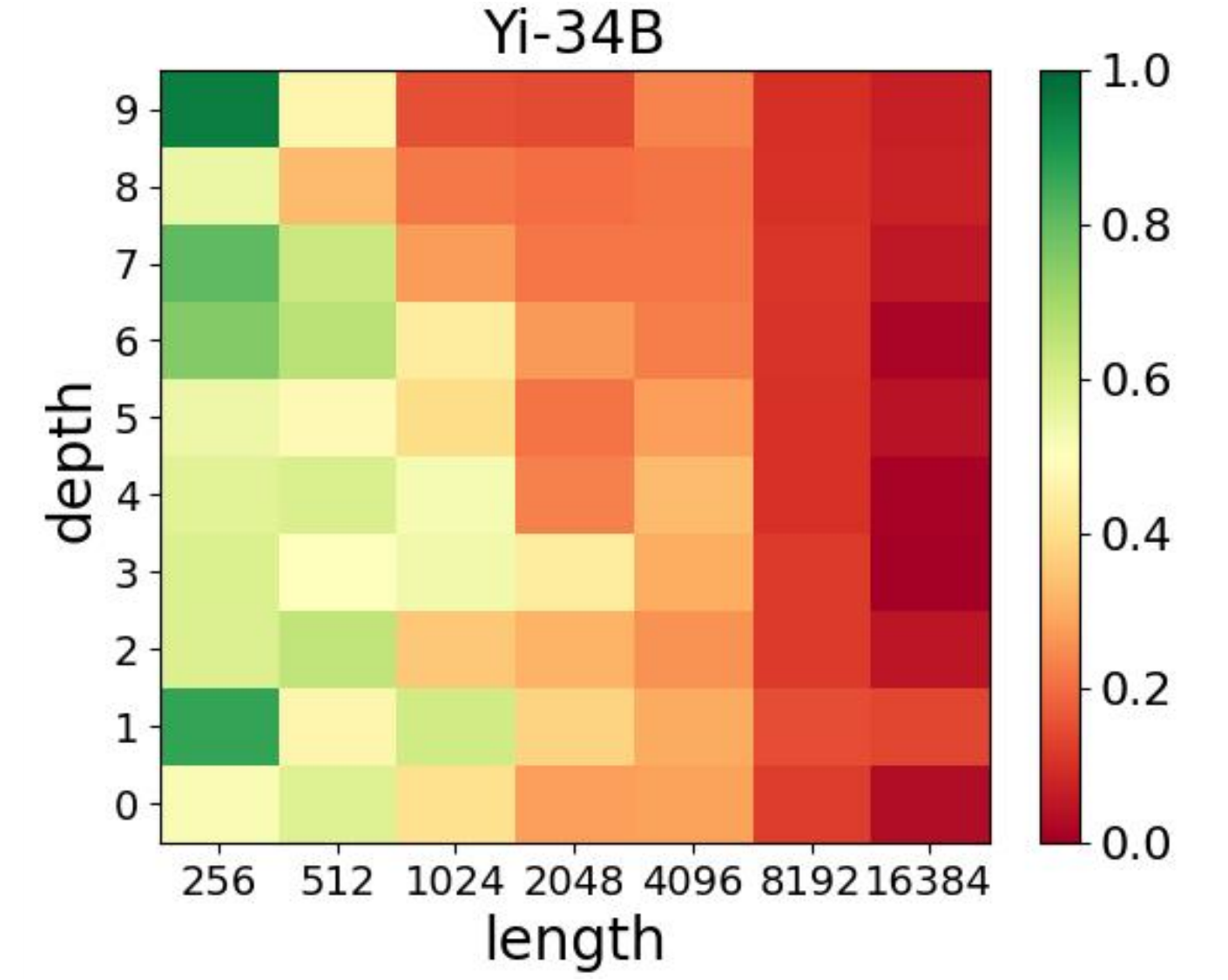}
        \label{fig:image6}
    \end{minipage}
    \caption{The impact of positional distribution and overall length of the questions on model performance is illustrated, with GPT series models on the left and Yi series models on the right. The horizontal axis represents the total length of the questions, while the vertical axis indicates the position of incorrect answers within the questions, where 0 denotes the beginning and 9 denotes the end. The greener the color, the higher the model's recognition accuracy at that position.}
    \label{fig:hotmap}
\end{figure}

Specifically, We analyze the model's sensitivity to position when dealing with lengthy key information by controlling the depth of incorrect questions within papers of various lengths and recording the model's accuracy at those positions. The results are shown in \autoref{fig:hotmap}.

Through the above experiment, we can clearly observe the performance of models when faced with incorrect answers at different depths and lengths of test papers. Using the GPT series and Yi series as examples, both exhibit the same characteristics: the models demonstrate better recognition ability at the beginning of the test paper for the same length, with performance gradually declining as depth increases. The longer the overall length of the test paper at the same depth, the worse the performance. This is generally in line with our expectations. However, it is noteworthy that, unlike needle-in-a-haystack or other long-text benchmarks, \bench only requires length of 16k to challenge the current state-of-the-art GPT-4 series models, while conventional closed-source models (such as the Yi series models) only require 8k to basically reach their performance limit. More depth-length two-dimensional diagrams for additional models can be found in \autoref{fig:hotmap_detailed} in the appendix.

\begin{table*}[htbp]
\centering
\small
\resizebox{\linewidth}{!}{%
\begin{tabularx}{\textwidth}{@{} l *{8}{X} c @{}} % l 表示左对齐的固定列，*{6}{X}表示六个自适应宽度的列，c表示一个中心对齐的固定列
% \begin{tabular}{@{}lccccccc@{}}
\toprule
\textbf{Model} & \textbf{Size} &\textbf{Support}  & \textbf{256} & \textbf{512} & \textbf{1024} & \textbf{2048} & \textbf{4096} & \textbf{8192} & \textbf{16384} \\ 
\midrule

GPT-4o & \faLock{} & 128k & 74.89& 74.62 & 70.43  & 66.98 &  61.82  &  59.07 &  54.30   \\ 
GPT-4-Turbo & \faLock{} & 128k & 71.62 & 67.47 & 61.11 & 59.36 & 56.62 & 51.01 & 44.91 \\
GPT-3.5-Turbo & \faLock{} & 16K & 53.03 & 44.28 & 40.16 & 33.80 & 26.55 & 19.81 & 14.03 \\ 
ERNIE-Speed & \faLock{} & 8k & 51.29 & 40.38 & 31.86 & 25.90 & 16.24 & 2.61 & - \\ 
Qwen-Long & \faLock{} & 10M & 64.85 & 56.32 & 44.46 & 36.11 & 29.05 & 23.16 & 15.44 \\ 
Deepseek-Chat & \faLock{} & 32k & 67.71    & 63.13 &  56.99 & 51.04 & 46.78 & 42.00 & 37.25 \\
Yi-Large-Turbo & \faLock{} &  200k & 60.05 & 53.78 & 42.11 & 36.09 & 27.20 & 21.66 & 10.75 \\ 
Yi-Spark & \faLock{} &  200k & 48.24 & 40.48 & 34.35 & 26.37 & 19.04 & 8.34 & 1.01 \\ 
GLM-4 & \faLock{} & 128K & 60.25 & 54.11 & 49.79 & 46.05 & 41.32 & 37.88 & 34.15 \\ 
[2pt]\hdashline\\[-8pt]
MAP-Neo-Ins-v0.1 & 7B & 8k & 39.34   & 33.71  &  28.45  &  23.82   & 16.04  &  6.83 & - \\
Deepseek-LLM-Chat & 7B & 4k & 17.70 & 12.93 & 8.42 & 7.47 & 1.16 & - & - \\ 
Qwen1.5-7B-Chat  & 7B & 32k & 25.93 & 24.71 & 20.81 & 14.71 & 12.35 & 11.05 & 10.69 \\
Llama3-8B-Ins  & 8B & 8k & 46.84 & 40.97 & 37.41 & 27.46 & 21.11 & 16.08 & - \\  
[2pt]\hdashline\\[-8pt]
LongChat-13B  & 13B & 16k & 28.19 & 25.13 & 23.00 & 19.27 & 15.32 & 9.22 & 6.45 \\ 
Baichuan2-13B-Chat  & 13B & 4k & 31.28 & 26.33 & 23.70 & 20.87 & 0 & - & - \\
\bottomrule
\end{tabularx}
}
\caption{The result of Local Instruction \& Single Task. ``Ins'' indicates ``Instruct''}
\label{tab:local single instructions}
\end{table*}

\subsection{Effect of the Density of Key Information}

\begin{figure}[t]
    \centering
    \includegraphics[width=0.45\textwidth]{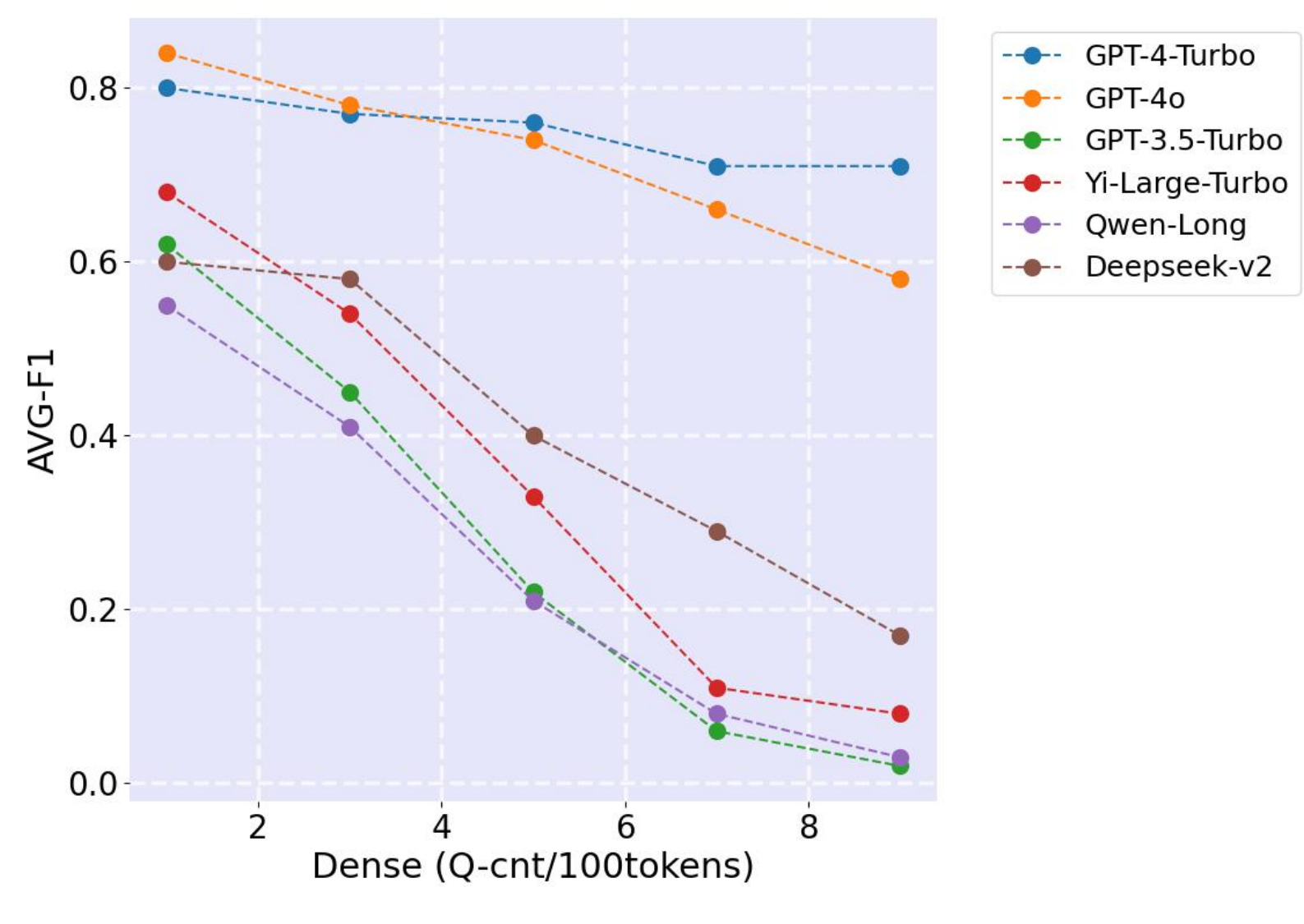}
    \caption{The impact of question density on model performance. The horizontal axis represents the number of questions per 100 tokens in the paper; the higher this value, the more questions are contained within the same total length. The vertical axis represents the model's score at that question density.}
    \label{fig:question dense}
    \vspace{-0.2cm}
\end{figure}

In our current setup, we concatenate questions of the same type to form a test paper, maintaining a constant length. However, due to variations in the lengths of the questions, the number of questions in papers of the same length varies. Therefore, we analyze the accuracy of multiple models with varying numbers of questions (i.e., question density) under the same test length, aiming to explore the impact of key information density on LLMs. We track the scoring results of test papers with different question densities and plot a graph of scores varying with question density.

As illustrated in \autoref{fig:question dense}, it is evident that most models exhibit a performance degradation as the number of questions increases under the same test paper length. 
This includes high-scoring models such as Qwen. 
This demonstrates most models are more sensitive to the number of questions when the total length is the same.

This also indicates that besides the length of the sample, the density of key information is also an important factor affecting performance. 
However, GPT-4-turbo and GPT-4o maintain good performance even at higher question densities, which to some extent indicates their robustness to changes in key information density. They can still maintain high-density information processing capabilities in long contexts.

\begin{table}[hbp]
\centering
\small
\begin{tabularx}{0.48\textwidth}{Xc}
\toprule
\textbf{Model} & \textbf{Accuracy} \\
\midrule
GPT-4-Turbo & 95.7 \\
GPT-3.5-Turbo & 87.0 \\
Deepseek-v1 & 88.2 \\
Yi-Large-Turbo & 87.8 \\
Llama3-8B-Instruct & 81.0 \\
ERNIE-Speed & 81.5 \\
GLM-4 & 90.1 \\
Qwen-Long & 83.6 \\
\bottomrule
\end{tabularx}
\caption{Directly present the QA of a single question to the model and have it determine whether the answer is correct, controlling the ratio of true labels to be 1:1. The accuracy of the LLMs is then evaluated.}
\label{tab:question acc}
    \vspace{-0.2cm}
\end{table}

\subsection{Ablation Experiment of Single Question}

\bench aims to assess the ability of LLMs to maintain focus over longer key information by identifying incorrect question numbers through a full-text review. However, LLMs might make recognition errors not due to its lack of contextual capabilities but because it inherently lacks the ability to answer questions. Therefore, we set up this ablation experiment to ensure a true evaluation of the long-context abilities of LLMs. We conduct experiments using single questions as test papers for several models. The results, as shown in \autoref{tab:question acc}, indicate that the accuracy of the models can generally exceed 80\%, with some, like GPT-4-turbo, reaching up to 95.7\%. These results suggest that \bench provides a reliable evaluation of the actual contextual length performance of LLMs.

\subsection{Effect of Different Task Types}
\begin{figure}[htbp]
  \centering
    \centering
    \includegraphics[width=1.0\linewidth]{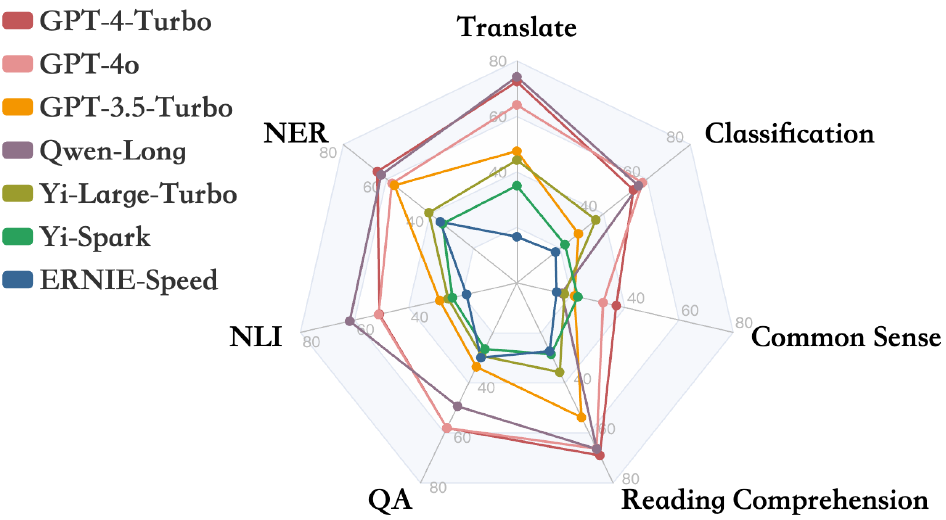}
  \hfill
    \centering
  \caption{The average performance of the model across different task types.}
  \label{fig:taskType}
    \vspace{-0.2cm}
\end{figure}

We also explore the performance of LLMs when using test papers composed of different task types., as shown in \autoref{fig:taskType}.

A notable trend is that the model generally performs poorly in the common sense category but achieves higher scores in tasks such as NER and Reading Comprehension. By analyzing the actual responses and the questions, we find that despite our efforts to standardize the format of questions within the same test paper, there are still significant distribution disparities among different task types. The questions in the common sense category tend to vary widely within the same test paper, whereas tasks like NER and Reading Comprehension have almost consistent formats, which can stimulate the model's in-context learning ability. When the variation is large, the advantage of in-context learning is almost nonexistent, leading to the model's generally poor performance.

%\subsection{Impact of Supervised Fine-tuning on \bench}

%To preliminarily explore the potential enhancement of LLMs' multi-hop compositional reasoning abilities, we conducted SFT training using 3000 training samples of the same paradigm, with Llama3-8B-Instruct as the base model. We then evaluated the results on our benchmark, as shown in \autoref{tab:local mixed instructions}.

%We can observe that models generally score higher under the mix-taskType mode compared to regular tests. This is because the questions from different taskTypes have unique instructions, necessitating separate instructions for each output. Consequently, the dependency on instructions becomes more pronounced, resulting in higher scores. In contrast, when comparing results in \autoref{tab:local single instructions}with the same local single instruction, it is evident that questions from different taskTypes are more challenging under the same instruction structure. This indicates that most models still have a certain advantage in in-context learning paradigm tasks.

\section{Conclusion}
In this paper, we introduce \bench Bench, a benchmark designed to evaluate the long-context processing capabilities of LLMs by assessing their true reading window length. \bench consists of over 1400 task types with varying lengths and controlled distributions of incorrect answers. Our findings highlight that models perform better with closer instruction distances, degrade with increasing length and density of key information, and exhibit varied performance across different task types. Notably, GPT-4-turbo and GPT-4o demonstrate robustness to high-density information. \bench provides a reliable framework for assessing LLMs' proficiency in understanding extensive sequences, offering valuable insights for future research and development.

\section*{Limitations}
Despite the comprehensive evaluation framework proposed by \bench, there are several limitations to our study. The test set in \bench is constructed using synthetic data, which, while controlled and consistent, may not entirely reflect the complexity and variability of natural language inputs. Additionally, our evaluation methodology emphasizes identifying incorrect answers within long texts, which may not cover other critical aspects of LLM performance such as creativity, user engagement, and adaptability to unforeseen inputs.

\section*{Ethics Statement}
This research adheres to ethical guidelines for AI development. We aim to enhance the capabilities of LLMs while acknowledging potential risks such as bias, misuse, and privacy concerns. To mitigate these, we advocate for transparency, rigorous bias testing, robust security measures, and human oversight in AI applications. Our goal is to contribute positively to the field and to encourage responsible AI development and deployment.

\bibliography{custom}

\appendix
\onecolumn
\clearpage % 强制换页

\section{Example Prompt for GIST with a Length of 256 Tokens}
\begin{table*}[htbp]
\centering
\resizebox{\linewidth}{!}{%
\begin{tabularx}{\textwidth}{@{}X@{}}
\toprule
\textbf{Prompt Example (256)}\\
\midrule
You are an excellent grading teacher, and you will be given many completed questions. The requirements for these questions are as follows: \\

\textcolor{red}{\texttt{[In this task, you need to answer the given multiple-choice question on the physics. Classify your answers into 'a', 'b', 'c', 'd', and 'e'.]}} \\

Please identify all the questions that have been answered incorrectly according to the requirements.\\

\textcolor{blue}{\texttt{[0:Problem: how many seconds does sandy take to cover a distance of 700 meters , if sandy runs at a speed of 18 km / hr?
Options: a.100 , b.120 , c.140 , d.160 , e.180 Answer: d]}} \\ 
\textcolor{blue}{\texttt{[1:Problem: a train 300 m long takes 9 sec to cross a man walking at 3 kmph in a direction opposite to that of the train . find the speed of the train? 
Options: a.100 kmph , b.90 kmph , c.120 kmph , d.117 kmph , e.25 kmph Answer: d]
}}\\ 
\textcolor{blue}{\texttt{[2:Problem: a 300 m long train crosses a platform in 39 sec while it crosses a signal pole in 9 sec . what is the length of the platform ? 
Options: a.389m , b.350m , c.289m , d.799m , e.1000m Answer: e]}} \\ 
The above text contains a description of the task for this group, as well as many question-answer pairs. Each enclosed in brackets. You are requested to identify which answer(s) are incorrect and to directly provide the question number(s) of the incorrect answer(s) enclosed in brackets. Please note, I need the numbers of the questions that are incorrect, do not provide the numbers of the questions with correct answers. Additionally, ensure to output the question numbers in the specified standard format.\\
Answer:\\
\bottomrule
\end{tabularx}
}
\caption{An example of a complete prompt under the length of 256 tokens.}
\label{tab:example_prompt}
\end{table*}
\FloatBarrier
% \clearpage % 强制换页
\section{HotMap of More Models}
\label{more_hotmap}
The 2D length-depth distribution map for more models. The horizontal axis represents the total length of the samples, while the vertical axis indicates the position of incorrect answers within the questions, where 0 denotes the beginning and 9 denotes the end. The greener the color, the higher the model's recognition accuracy at that position.

\begin{figure*}[h] % 使用figure*环境
    \centering
    \begin{minipage}{0.48\textwidth}
        \centering
        \includegraphics[width=\textwidth]{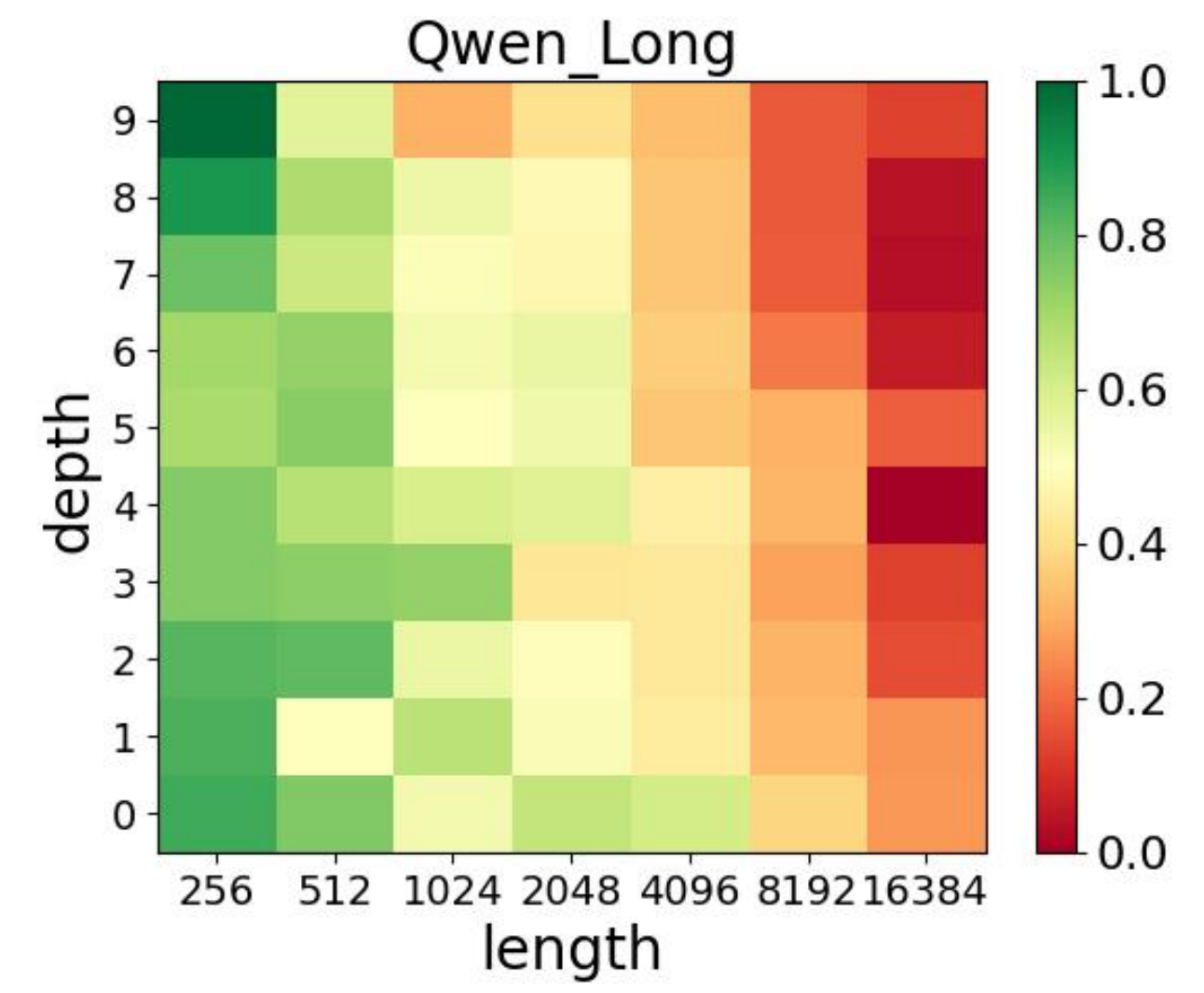}
        \label{fig:image1}
    \end{minipage}
    \hfill
    \begin{minipage}{0.48\textwidth}
        \centering
        \includegraphics[width=\textwidth]{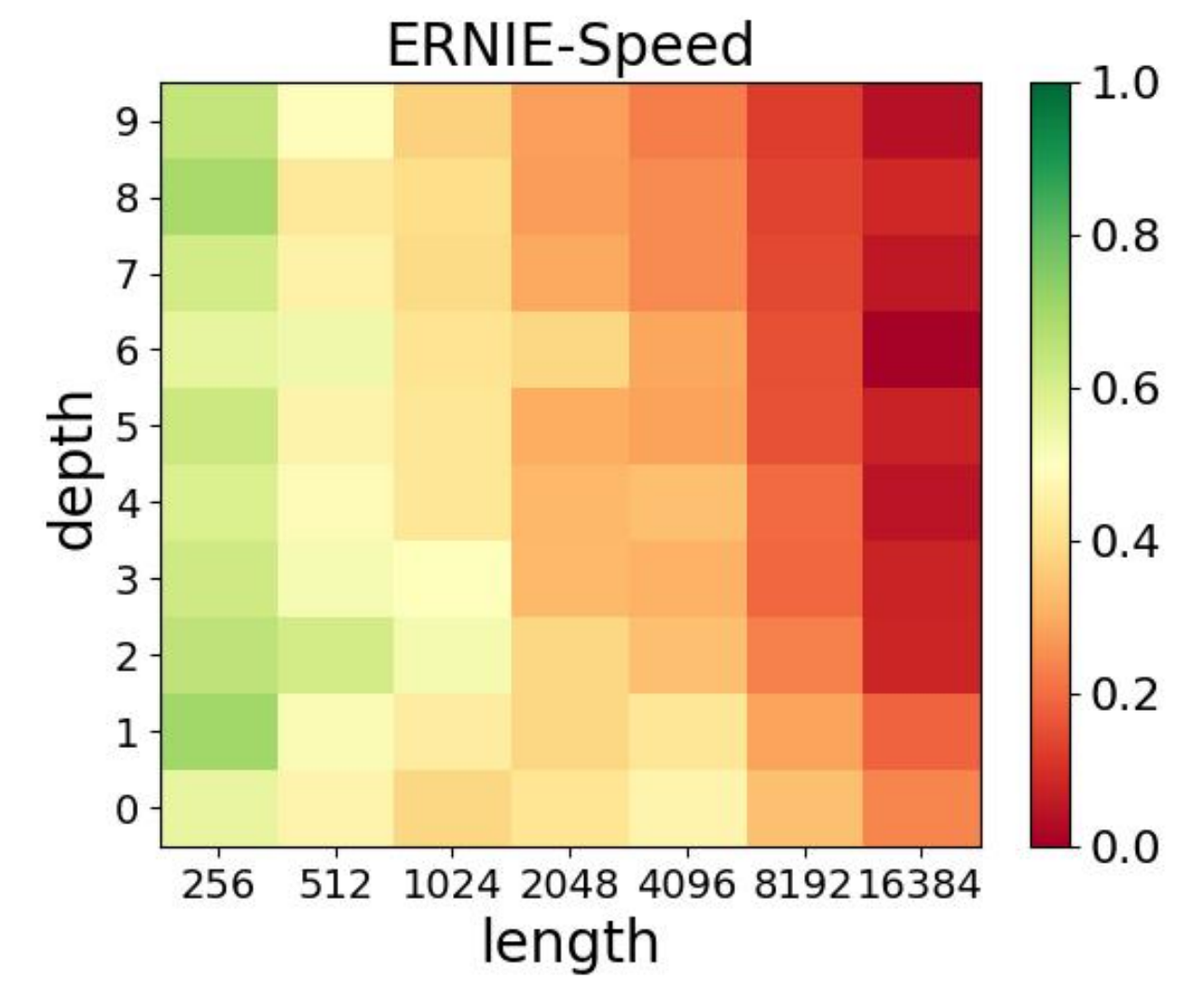}
        \label{fig:image2}
    \end{minipage}
    \hfill
\end{figure*}
\begin{figure*}[h] % 使用figure*环境
    \begin{minipage}{0.32\textwidth}
        \centering
        \includegraphics[width=\textwidth]{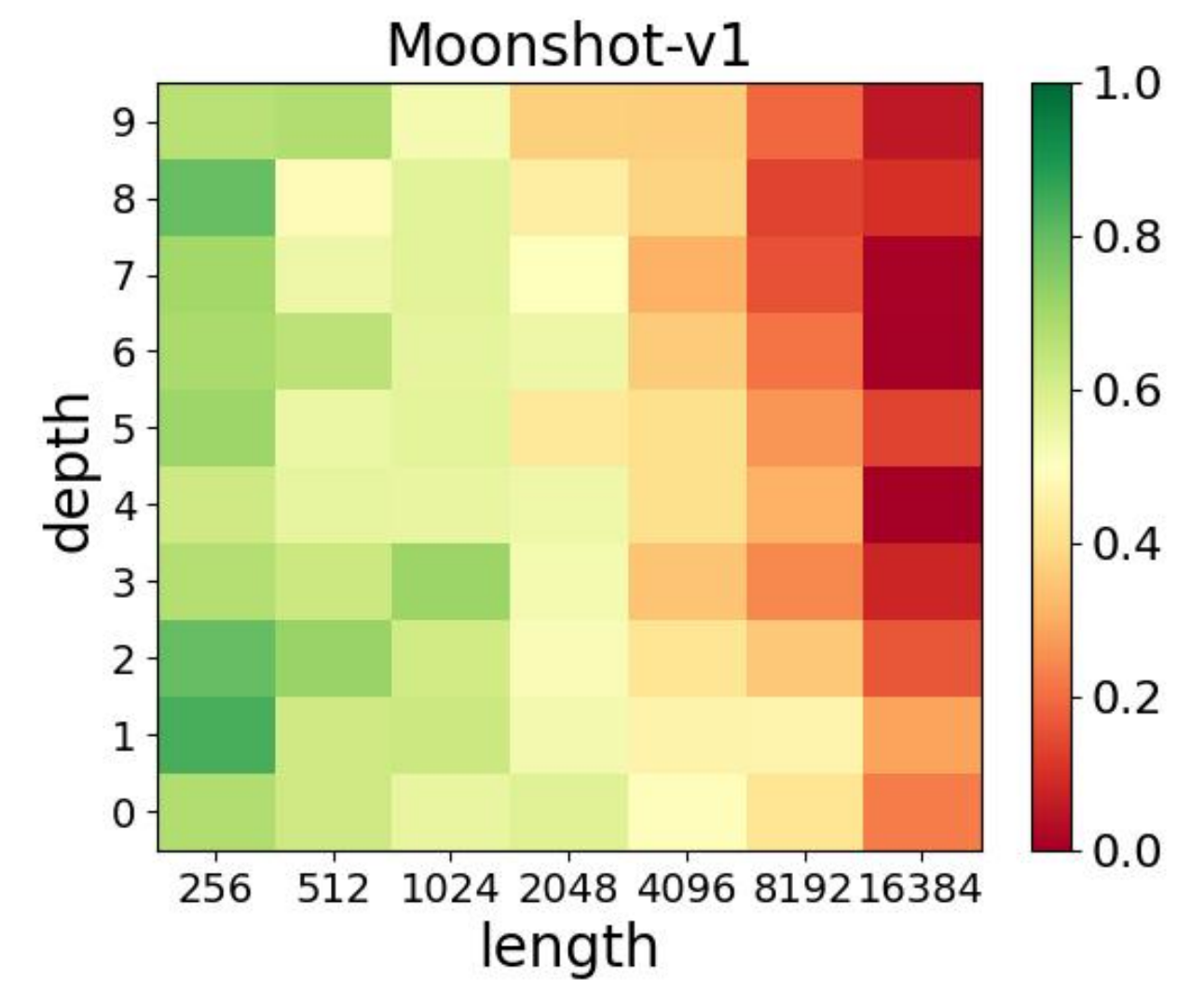}
        \label{fig:image3}
    \end{minipage}
    \hfill
    \begin{minipage}{0.32\textwidth}
        \centering
        \includegraphics[width=\textwidth]{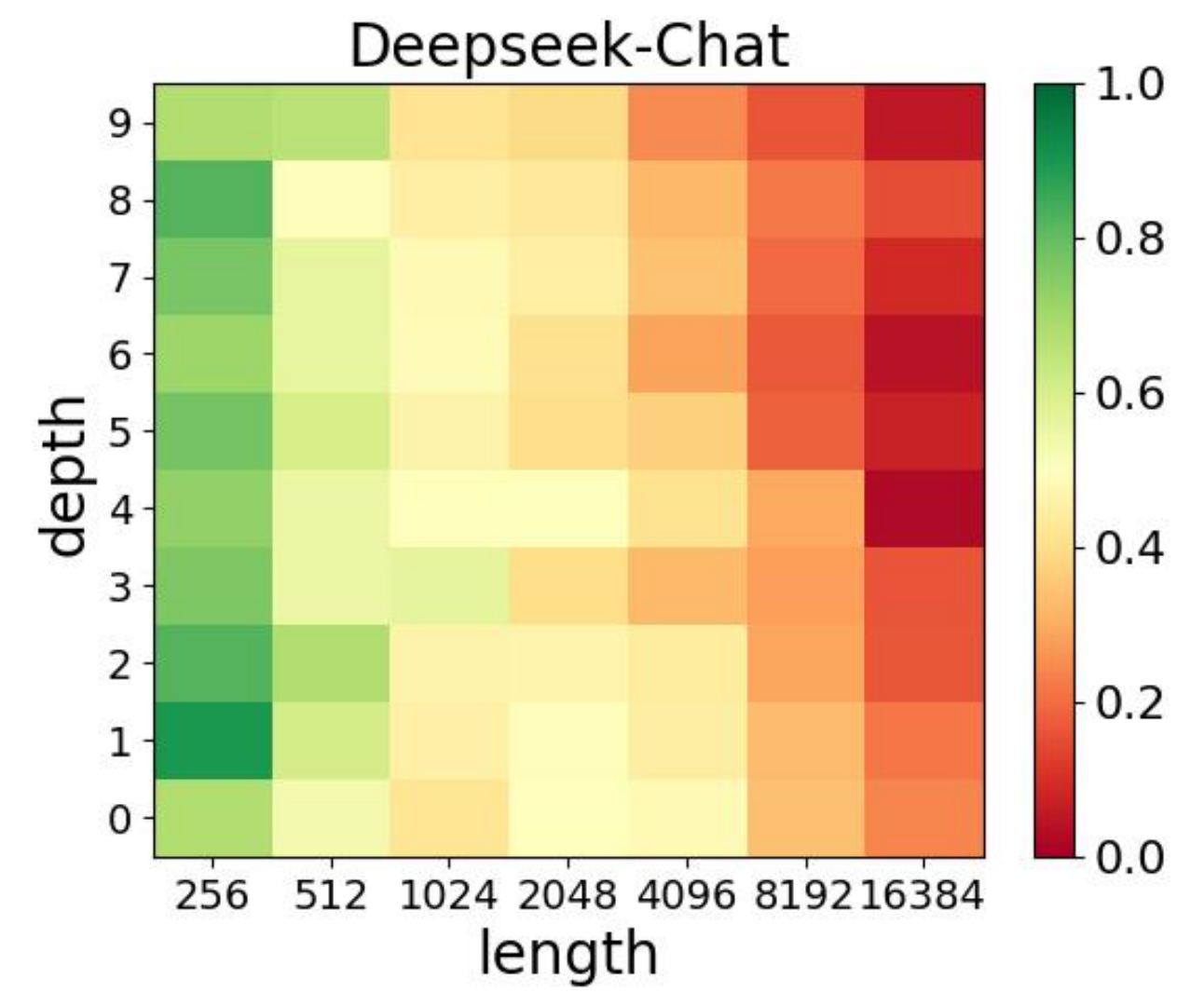}
        \label{fig:image4}
    \end{minipage}
    \hfill
    \begin{minipage}{0.32\textwidth}
        \centering
        \includegraphics[width=\textwidth]{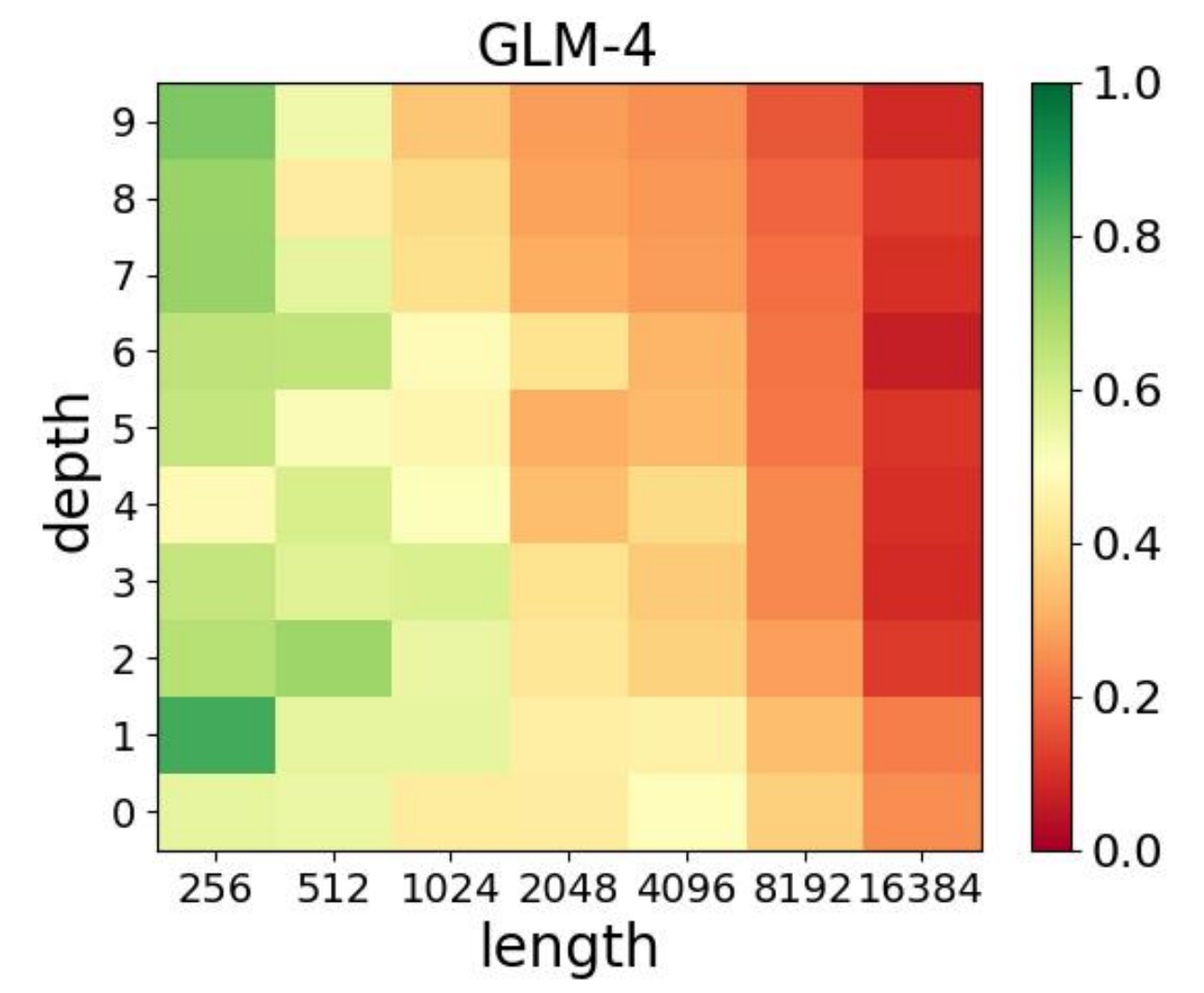}
        \label{fig:image6}
    \end{minipage}
    \caption{Expanding the Model's Heatmap.}
    \label{fig:hotmap_detailed}
\end{figure*}
\FloatBarrier

\section{Prompt Template}

\subsection{Template of GIST}
\begin{figure*}[h!]
\centering
\resizebox{1\textwidth}{!}{ % 调整表格宽度为页面宽度的0.9
\begin{promptbox}[Prompt Template]{lightred}
You are an excellent grading teacher, and you will be given many completed questions. The requirements for these questions are as follows: \\
\textcolor{red}{\texttt{[The Task Description]}} \\
Please identify all the questions that have been answered incorrectly according to the requirements.\\
\textcolor{red}{\texttt{[Test Paper]}} \\ 
The above text contains a description of the task for this group, as well as many question-answer pairs. Each enclosed in brackets. You are requested to identify which answer(s) are incorrect and to directly provide the question number(s) of the incorrect answer(s) enclosed in brackets. Please note, I need the numbers of the questions that are incorrect, do not provide the numbers of the questions with correct answers. Additionally, ensure to output the question numbers in the specified standard format.\\
Answer:
\end{promptbox}
} % 结束resizebox
\captionof{figure}{The prompt template used for GIST in \bench.}
\label{fig:prompt}
\end{figure*}

\FloatBarrier

\subsection{Template of LIST}
\begin{figure*}[h!]
\centering
\resizebox{1\textwidth}{!}{ % 调整表格宽度为页面宽度的0.9
\begin{minipage}{\textwidth} % 使用minipage保持内部格式
\begin{promptbox}[Prompt Template]{lightred}
You are an excellent grading teacher, and you will be given many completed questions.  \\
\textcolor{red}{\texttt{[The Task Description,Question 1]}} \\ 
\textcolor{red}{\texttt{[The Task Description,Question 2]}} \\ 
\textcolor{red}{\texttt{[The Task Description,Question 3]}} \\ 
\textcolor{red}{\texttt{[The Task Description,     ........     ]}} \\ 
\textcolor{red}{\texttt{[The Task Description,Question N]}} \\ 
The above text contains a description of the task for this group, as well as many question-answer pairs. Each enclosed in brackets. You are requested to identify which answer(s) are incorrect and to directly provide the question number(s) of the incorrect answer(s) enclosed in brackets. Please note, I need the numbers of the questions that are incorrect, do not provide the numbers of the questions with correct answers. Additionally, ensure to output the question numbers in the specified standard format.\\
Answer:
\end{promptbox}
\end{minipage}
} % 结束resizebox
\captionof{figure}{The prompt template used for LIST in \bench.}
\label{fig:prompt_local}
\end{figure*}
\FloatBarrier

\end{document}